# BSM loss A superior way in modeling aleatory uncertainty of fine-grained classification

Shuang Ge[1], Kehong Yuan[1], Maokun Han[2], Desheng Sun[3], Huabin Zhang[4], Qiongyu, Ye[5]
[1] Graduate School at Shenzhen, Tsinghua University, Shenzhen, China
2 Ping An Technology, Shenzhen, China
[3] Shenzhen Hospital of Peking University, Shenzhen, China.
[4] Beijing Tsinghua Changgung Hospital, Tsinghua University, Beijing, China
[5] Shenzhen City Baoan District Women's and Children's Hospital, Shenzhen, China.

**Abstract.** Artificial intelligence(AI)-assisted method had received much attention in the risk field such as disease diagnosis. Different from the classification of disease types, it is a fine-grained task to classify the medical images as benign or malignant. However, most research only focuses on improving the diagnostic accuracy and ignores the evaluation of model reliability, which limits its clinical application. For clinical practice, calibration presents major challenges in the low-data regime extremely for over-parametrized models and inherent noises. In particular, we discovered that modeling data-dependent uncertainty is more conducive to confidence calibrations. Compared with test-time augmentation(TTA), we proposed a modified Bootstrapping loss(BS loss) function with Mixup data augmentation strategy that can better calibrate predictive uncertainty and capture data distribution transformation without additional inference time. Our experiments indicated that BS loss with Mixup(BSM) model can halve the Expected Calibration Error(ECE) compared to standard data augmentation, deep ensemble and MC dropout. The correlation between uncertainty and similarity of in-domain data is up to -0.4428 under the BSM model. Additionally, the BSM model is able to perceive the semantic distance of out-of-domain data, demonstrating high potential in real-world clinical practice.

## 1 Introduction

Deep learning models have achieved enormous advances in a variety of research fields, as for medical image analysis and computer aided diagnosis[1-4]. In many situations, over-parametrized models are easily vulnerable to calibration issues, especially in the case of insufficient data and inherent noises (including label noises, instruments and equipment noises, and manual acquisition noises, etc.)[5]. Moreover, the widely used cross entropy loss with *softmax* activation causes disaster "over-confident" predictions[6]. It is essential to improve the accuracy of model, but the neural network "know what they don't know" has become increasingly important in clinical practice. While many types of research focused on improving the accuracy of the predictions, the uncertainty calibration (UC) of the predictions mostly remains unnoticed.

To overcome calibration deterioration, aleatoric uncertainty and epistemic uncertainty have been proposed[7–9] to quantify uncertainty, such that uncertain referrals can be transferred to human experts. Epistemic uncertainty is also known as model uncertainty and can be explained away given enough data. Aleatoric uncertainty captures noise inherent in the observations, corresponding to the input-dependent (data) uncertainty. Aleatoric uncertainty is irreducible, even with more training data[8,10].

Bayesian neural networks (BNNs) replace the weight parameters of the deterministic network with distributions over these parameters[11, 12]. They are easy to formulate, but difficult to perform inference in[8]. In recent years, research have shown an increased interest in modeling uncertainty, such as ensemble approaches(Averaging, Voting[13], Bagging[14], Gradient Boosting[15]) , and Monte Carlo (MC) Dropout[16]. They all performed exceedingly well on large visual datasets, but sometimes badly on representation learning tasks in low-data regimes[7]. Ensemble is also typically costly with many networks trained and inferred in parallel.

For medical image processing, due to the divergence in medical equipment and acquisition process, we can hardly reduce the inherent noises by adding more training data[5, 8]. Therefore, we believe that modeling the aleatoric uncertainty may be more imperative in low-data clinical practice. Test-time augmentation is a commonly used method to modeling data uncertainty during testing, and it costs much more inference time than the single forward network. Ayhan, M. S. et al.[17] explored the uncertainty estimation and the calibration of TTA in diabetic retinopathy detection and showed well calibrated measures of uncertainty in clinical practice.

Another essential factor that is needed for a proper uncertainty-aware algorithm is being distanced aware and domain-aware. It means that the algorithm distinguishes when the test data are far from those it was trained on[18]. However, the representation of distance between two datasets were not discussed in many works.

This paper studies the automatic diagnosis of the benign and malignant datasets, including the Skin Cancer data set and the clinical Breast Ultrasound data set. Based on empirical observations, we studied the trade-offs between the data-dependent uncertainty and model-dependent uncertainty. Deep Ensemble, TTA algorithms are applied to the Skin data set separately and simultaneously, and MC Dropout is also discussed because it is also one of the classic methods for modeling model uncertainty. We also compare the effects of different percentages of training data on model performance in a data-dependent way. In addition, we proposed a new method to modeling data-dependent uncertainty(both label noise and data noise) during training, using commonly data augmentation, TTA as the comparison. Besides, we consider the popular metrics for evaluating predictive uncertainty, e.g. ECE, NLL, and Brier score[7, 19]. Calibration can reflect the noise modeling of the model, but the perception of semantic distance and distributed transformation are equally important. We compared the sensitivity to data distribution and domain transformation for data-dependent uncertainty. To the best of the author's knowledge, we are the first to perform uncertainty-aware on both in-distribution and out-of-distribution data on small datasets of medical images.

The rest of this paper is organized as follows. Section 2 illustrates the uncertainty-aware algorithms proposed in this study. The Experimental details, including dataset introduction, parameters setting, uncertainty quantification and metrics, are described in Section 3. In Section 4, we showed the result of different methods, in addition to our Empirical Observations about deep ensemble, TTA and Mc dropout on the Skin dataset, we focused on comparing different data-dependent uncertainties calibration and perception on different dataset under different noise levels.

## 2 Method

### 2.1 Mixup

Despite being robust to small amounts of label noise, the loss function of convolutional neural network is easy to overfit the noise label, especially for the relatively small amount of medical images[20]. *Mixup*[21] has recently demonstrated **outstanding** robustness against label noise without explicitly modeling. For two feature-target vectors $(x_i, y_i)$ and $(x_j, y_j)$ drawn at random from the training data, the loss function can be defined as

$$\mu(x', y' \mid x_i, y_i) = \frac{1}{N}\sum_{j}^{N} \mathop{\mathrm{E}}_{\gamma}\left[\delta(x', y')\right] \quad (1)$$

Where $x' = \gamma x_i + (1-\gamma)x_j$ and $y' = \gamma y_i + (1-\gamma)y_j$

A random coefficient $\gamma \in (0,1)$ drawn from a fixed mixing distribution often chosen as $Beta(\alpha, \beta)$ where $\alpha = \beta$.

### 2.2 Bootstrapping loss

We first consider BS loss[22] in modeling aleatoric uncertainty. Different from focal loss, cross-entropy loss(CE loss) and other empirical risk function, BS loss maintains a higher cross-entropy loss for noise samples during training[23]. The Beta Mixture Model (BMM) better approximates the loss distribution for decision boundaries that transform linearly from clean to noisy samples. The static BS loss adds weighted labels and predicted values to the standard cross-entropy loss that helps to correct the training objective function:

$$l_{BS} = -\sum_{i=1}^{N}((1-w_i)y_i + w_i z_i)^T \log(h_i) \quad (2)$$

where $y_i$ denotes the label and $w_i$ weights the model prediction label $z_i$ , $h_i$ denotes the prediction probability on the predict class of the model.

### 2.3 Bootstrapping loss with Mixup (BSM)

We propose to fuse *Mixup*(Eq. 1.) and hard bootstrapping(Eq. 4.) to implement a robust per-sample loss correction approach and provide a smoother estimation of uncertainty:

$$l_{BSM} = -\gamma\left[((1-w_i)y_i + w_i z_i)^T \log(h_i)\right] - (1-\gamma)\left[((1-w_j)y_j + w_j z_j)^T \log(h_j)\right] \quad (3)$$

The weights $w_i$ and $w_j$ that control the proportion of labels and predictions are inferred from beta mixture model (BMM) model, $w_i$ indicates $p(k = y_i | y_i)$, and $w_j$ indicates $p(k = y_j | y_j)$, $\gamma$ is the *Mixup* hyper-parameter.

Because the noise sample often shows higher cross-entropy loss than that of clean ones during training, we can infer from the loss value if a sample is more likely to be clean or noisy[23]. Supposed that the BMM model is estimated after each training epoch using the cross-entropy loss $l_i$ for each sample *i*. Based on $l_i$ ,We use posterior probability to denotes clean (noisy) classes. Therefore, clean samples rely on their ground-truth label $y_i$ ($(1-w_i)$ is large), while noisy ones let their loss being dominated by their class prediction $z_i$ or their predicted probabilities $h_i$ ( $w_i$ is large).

This method model the data-dependent uncertainty during training, which can save much inference time than modeling uncertainty during test time.

# 3 Experimental details

## 3.1 Datasets

We evaluated and compared methods using two collections of images:(1) Skin Cancer data set from Kaggle[24] which contains 1800 benign cases and 1497 malignant cases. (2) Breast ultrasound of benign and malignant data set: it is collected from clinical, including 2064 benign lesions and 2064 malignant lesions acquired from Philips, Mindray, Hitachi, GE, etc. Five sophisticated doctors labeled the data independently, each of whom also examined the data labeled by others. All datasets were divided into training sets and validation sets with a 4:1 ratio. All methods used the same backbone of the modified Resnet-18 in this paper(Fig. 1. backbone).

## 3.2 Parameters setting

The weights of networks are initialized with Kaiming initialization[25] and we use weight decay with λ= 5e-4. We trained this module from scratch, depending on the distribution gap between natural and medical images. Empirically, we proved it better than pretrained trick and other fine-tuning operations related to natural images. We trained the model using an Adam optimizer. The initial learning rate is 5e-4 and it decays exponentially with a rate of 0.95 by epoch. Meanwhile, the minibatch size is fixed to 32, which is not a sensitive parameter in the proposed architecture. The mentioned results in this paper are performed by the most accurate model on validation data, using early stopping with “max” mode and ‘patience’=20. The configuration of best performance is saved for inference. Our source code is based on PyTorch1.6 and all experiments are conducted on a server with two GeForce GTX 2080Ti GPUs.

**Averaging Ensemble.** We averaged the outputs of multiple modified Resnet-18 for ensemble as shown in Fig. 1. All models were randomly initialized as different submodels. We followed previous works[9, 26, 27] and used the entropy of the mean prediction of multiple models to estimate the prediction uncertainty as below:

$$u_i = -\sum_k \mu_i^k \log(\mu_i^k) \quad (4)$$

where $i$ indicates the sample index, and $k$ indicates the class number. $\mu_i = \frac{1}{M}\sum_{m=1}^{M} p_i^m$, where $p_i^m$ is the *softmax* output of every estimator. We used $M$ to represent the number of ensemble models with single model($M$=1) and ensemble models($M$=3, $M$=5, $M$=10) employed in the experiment.

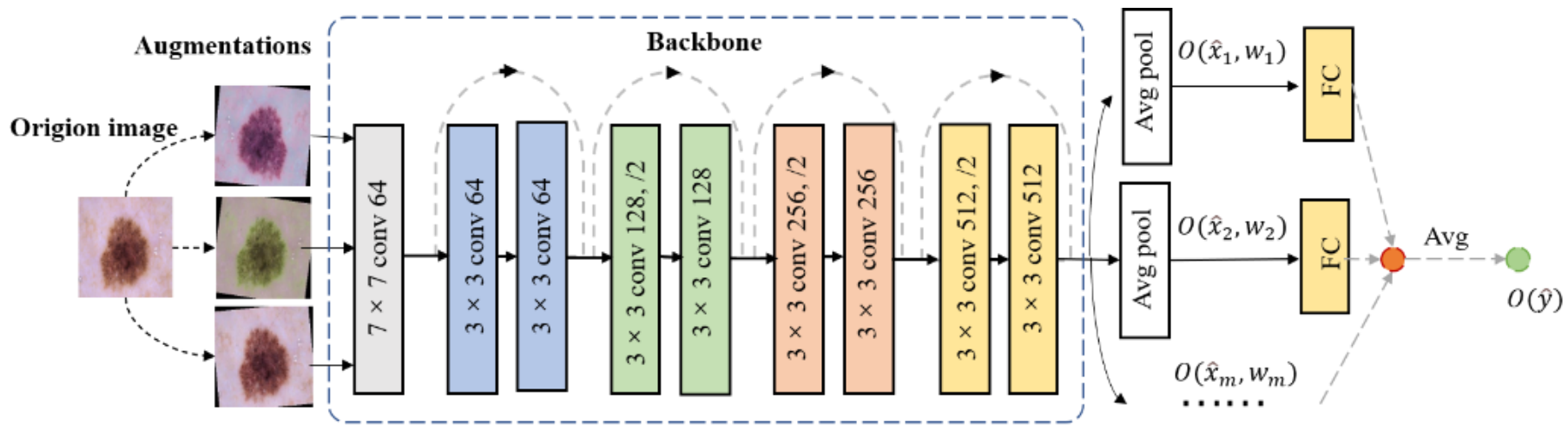

**Fig. 1.**The average of multiple identical networks' outputs was defined as the ensemble output in this paper.

**TTA.** We adopted image augmentation model following Wang, G. et al[28]. Assuming prior distributions over the parameters of transformations, we approximated the distribution of observations $p(x_0)$ as that of augmented images $p(x)$. For medical images, one should refrain from excessive transformations that might severely alter the structure in the data[17]. We mainly use the following operations in our data augmentation pipeline during both training and testing: (1) Color jitter. Brightness: $b \sim U(\phi_{11}, \phi_{12})$ where $\phi_{11} = -0.15$ and $\phi_{12} = 0.15$; Hue: $h \sim U(\phi_{21}, \phi_{22})$, where $\phi_{21} = -0.15$ and $\phi_{22} = 0.15$; Saturation: $s \sim U(\phi_{31}, \phi_{32})$, where $\phi_{31} = -0.15$ and $\phi_{32} = 0.15$; Contrast: $c \sim U(\phi_{41}, \phi_{42})$, where $\phi_{41} = -0.15$ and $\phi_{42} = 0.15$. (2) Geometric transformations. Horizontal flip is controlled by $f_h \sim Bern(\phi_5)$ where $\phi_5 = 0.5$; Translation offsets by pixels in 2D: $t_x \sim U(\phi_{61}, \phi_{62})$ and $t_y \sim U(\phi_{61}, \phi_{62})$ where $\phi_{61} = -22$ and $\phi_{62} = 22$; Rotation angle: $r \sim U(\phi_{71}, \phi_{72})$ where $\phi_{71} = -10$ and $\phi_{72} = 10$. (3) Resize: We resized all images to (224, 224) as the network inputs.

**Mc Dropout.** We followed Yarin Gal et al.[16]. The predictive mean and variance of the model can be written as follows:

$$E[y] \approx \frac{1}{T}\sum_{t=1}^{T} \hat{y}_t(x) \quad (5)$$

$$Var_q(y) \approx \tau^{-1} I_D + \frac{1}{T}\sum_{t=1}^{T} \hat{y}_t(x)^T \hat{y}_t(x) - E[y]^T E[y] \quad (6)$$

In the above equations, *x* is the input, $\hat{y}_t(x)$ is the corresponding output and $\tau$ is a constant value defined based on the model structure and T is the number of stochastic forward passes during the test time.

**Mixup.** Compared to Ensemble and TTA, *Mixup* is a data augmentation technique that implements only during training and it costs the least to be inferred during the test time. We try different $\alpha$ (Table 1 in Supplements) and finally choose $\alpha = 0.3$ for all experiments. This may simplify data augmentation, due to the less complex structure of medical data. We also find that high values of $\alpha = \beta$ may cause the deterioration in the calibration(e.g. $\alpha = \beta = 32$, $\gamma$ tends to be close to 0.5, see Table 1 in Supplements for details).

### 3.3 Uncertainty quantification

In this paper we used entropy for uncertainty estimation(Eq. 5)[18].

$$
\begin{aligned}
H_i(\hat{y}_i \mid x; \theta) &= -\sum_k p(\hat{y}_i = k \mid x_i; \theta) \log p(\hat{y}_i = k \mid x_i; \theta) \\
&= -\sum_k p_k \log p_k
\end{aligned}
\tag{7}
$$

where $p_k$ represents the prediction associated with class $k$. $\theta$ denotes the model's parameters and, $H_i$ represents the uncertainty predicted for $y_i$. For randomly initialized deep ensemble, TTA, and MC dropout, we all use the multi-times prediction mean as the final probability. For BSM and Mixup, we only use one single forward propagation to get the final result.

### 3.4 Metrics

**Expected Calibration Error(ECE).** The Expected Calibration Error (ECE) measures the divergence between empirical accuracy and prediction confidence. Given N predictions grouped into M bins. The equation is defined as

$$
ECE = \sum_{m=1}^{M} \frac{|B_m|}{N} \left| conf_m - acc_m \right| \tag{8}
$$

where $acc_m = \frac{1}{|B_m|} \sum_{i \in B_m} 1(\hat{y}(x_i) = y_i)$ and $conf_m = \frac{1}{|B_m|} \sum_{i \in B_m} \hat{p}(x_i)$. The $\hat{y}(x_i)$ and $y_i$ are the predicted and actual class labels, respectively. $B_m$ is the mth bin size, where we use 0.1 as a hyperparameter.

**Negative Log-likelihood(NLL).** This is an estimation of how well the model fits the prediction and the uncertain.

$$
NLL = -\frac{1}{n} \left[ \sum_{\{i \mid t_i = +1\}} \log p(y_i = +1 \mid x_i) + \sum_{\{i \mid t_i = -1\}} \log[1 - p(y_i = +1 \mid x_i)] \right] \tag{9}
$$

where $t_i$ is the target(for two classification task, +1 is the positive class while -1 represents the negative class) and the $y_i$ is the prediction class. So here we actually calculated the true class probability entropy(probability of prediction on label). It indicates how the NLL becomes infinite as the predictor becomes increasingly certain when test point was misclassified[19, 29].

**Brier score(BS).** It is another metric commonly used to evaluate the model calibration and penalize both the over-confidence and the under-confidence. It is defined as:

$$BS = K^{-1}\sum_{k=1}^{K}\left(t_k^* - p(y=k\mid x^*)\right)^2 \tag{10}$$

where $t_k^* = 1$ if $k = y^*$, and 0 otherwise[19, 30].

# 4 Result

## 4.1 Empirical Observations

**Classifier performance.** Fig. 2. shows the classifiers' performance of varying amounts of ensemble models on skin cancer dataset. We applied various degrees of TTA to the ensemble models, from 0~128(see Fig. 1. in Supplements). Here we discussed TTA and deep ensemble separately and simultaneously because we think the former is a data-dependent uncertainty prediction and the latter is about the model-dependent and we also want to explore the influence of the combination of the two on the model results. We discovered that the Area Under the Curve(AUC) generally declines as we expanded the number of integrated models. It implies that randomly initialized ensemble is not always beneficial and not always makes sense. Although there are many ways to ensemble, it does not work in all the cases, especially for small datasets. In addition, we trained models with two levels of augmentation, "tta0.5" and "tta0.9", which means that we transformed data with a 50% and 90% probability of color and geometric transformation respectively. We discovered that classifiers gained moderate improvement with the increasing TTA numbers while significant improvement for the increase of training augmentation proportion. Therefore, we supposed modeling data-dependent uncertainty during training may be more important in low-data regime.

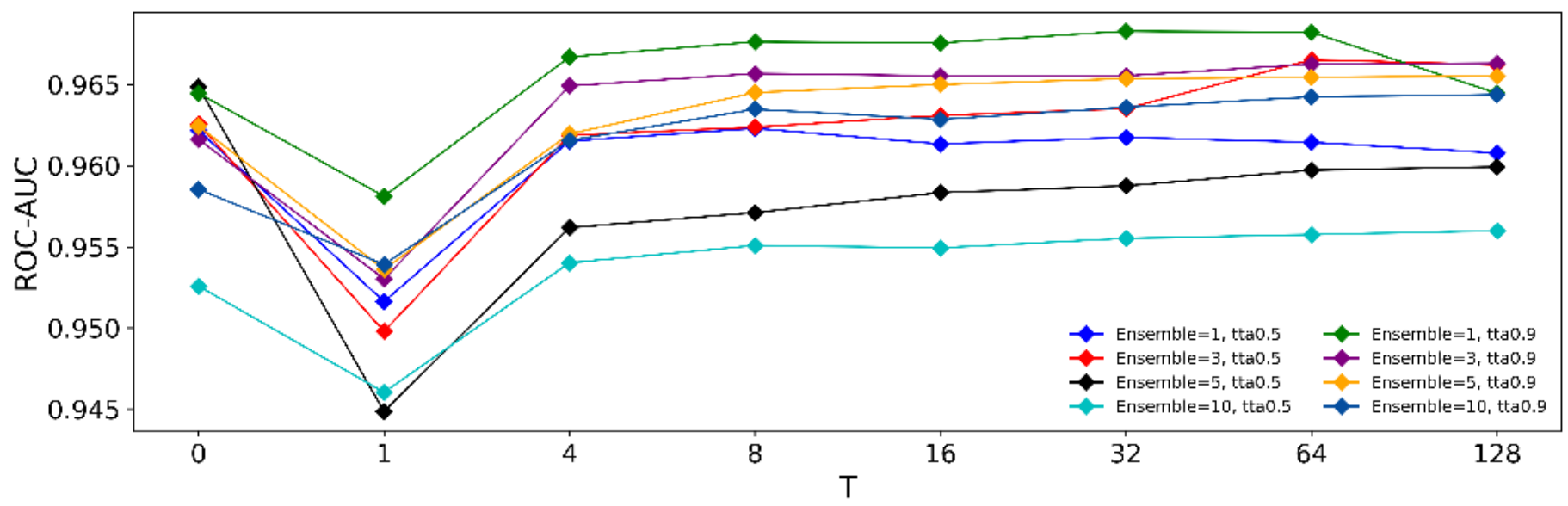

**Fig. 2. ROC-AUC vs. T (TTA times).** Four levels of ensemble(M=1, 3, 5, 10) with two degrees of training-time augmentation("tta0.5" and "tta0.9") and eight degrees of TTA(0, 1, 4, 8, 16, 32, 64, 128). "TTA0" means the original image was used for test.

Furthermore, we observed the calibration performance and the accuracy (classification threshold is set to 0.5) of the model. It shows that randomly initialized ensemble models are generally overconfident (Fig. 2. in Supplements) and gain no superior discriminator performance (NLL and brier, see Table 1.). Additionally, we discovered that TTA had a positive effect on reliability(ECE) but no significant effect on NLL and Brier score(Fig. 3. in Supplements). The results of MC dropout are shown in Table 2. As the number of forward propagation times increasing, the AUC, NLL, and Brier score improve while the ECE decline.

**Table 1.** Metrics of ensemble models(TTA:0) in skin cancer dataset

| | Ensemble=1 | Ensemble=3 | Ensemble=5 | Ensemble=10 |
|---|---|---|---|---|
| ROC-AUC ↑ | **0.9645** | 0.9616 | 0.9624 | 0.9586 |
| ECE ↓ | 0.0418 | **0.0340** | 0.0410 | 0.0406 |
| Brier score ↓ | 0.0742 | **0.0720** | 0.0743 | 0.0781 |
| NLL ↓ | **0.3952** | 0.4055 | 0.3961 | 0.4149 |

**Table 2.** Metrics of Mc Dropout(TTA:0) in skin cancer dataset

| | D=1 | D=4 | D=8 | D=16 | D=32 | D=64 | D=128 |
|---|---|---|---|---|---|---|---|
| ROC-AUC ↑ | 0.9660 | 0.9659 | 0.9665 | 0.9668 | 0.9668 | 0.9669 | **0.9671** |
| ECE ↓ | 0.0420 | **0.0381** | 0.0440 | 0.0534 | 0.0500 | 0.0500 | 0.0523 |
| Brier score ↓ | 0.0775 | 0.0766 | 0.0765 | 0.0773 | 0.0767 | **0.0763** | 0.0764 |
| NLL ↓ | 0.3807 | 0.3739 | 0.3694 | 0.3699 | 0.3682 | 0.3660 | **0.3657** |

**Decision referral.** We analyzed how the number of ensembles and TTA modify the referral of decisions. Results of the study are presented in Fig. 3(a) and Fig. 3(b). It indicates a positive effect in TTA while no or negative effects in ensemble. So we further improved our hypothesis that modeling aleatoric uncertainty may be more effective in low-data regimes.

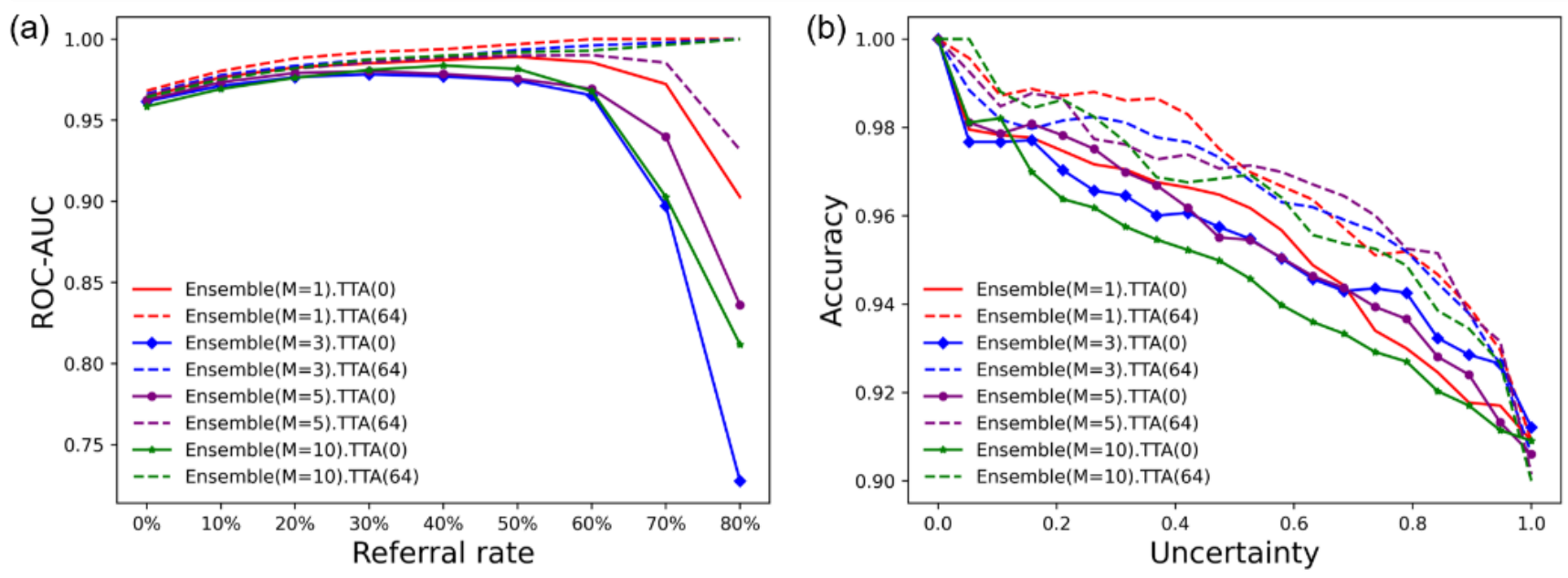

**Fig.3. Decision referral.** (a) Variations in performance via uncertainty decision referral. The classification performance of the remaining samples was expected to improve when the samples with large sorted uncertainty were rejected by a certain percentage. (b) Reduction in accuracy via uncertainty thresholds. When the uncertainty threshold of the excluded samples rises, the classification accuracy of the classifier will decline. TTA(64) is better than TTA(0) while ensemble is not always beneficial.

### 4.2 Calibrations and noise robustness

For aleatoric uncertainty, we evaluated the robustness of TTA, *Mixup* and BSM to inherent noise and artificial noise by manually adding 0% label noise and 20% label noise during training. The experimental results reveal that (1) TTA gains a better calibration, lower NLL and brier score than models without augmentation. (2) *Mixup* provides more reliable(ECE) predictions and lower NLL than non-*Mixup* model. (3) *Mixup* combined with BS loss is profitable to AUC, NLL, and brier score than *Mixup* only, while the calibration is comparable to TTA. (4) BSCE loss only outperforms other methods on breast ultrasound dataset. This may be due to the heavier noise involved in the ultrasound data. (Table 3) (5) Among all the data-dependent uncertainty, only BSM outperforming both MC DROPOUT and Deep ensemble methods in all three metrics (ECE, NLL and Brier score) on Skin dataset while MC dropout performs best on the AUC (0.9671 in Table 2).

**Table 3.** Calibration(ECE) and performance of different methods

| | No Mixup, CE loss | No Mixup, CE loss, TTA:64 | Mixup=0.3, CE loss | Mixup=0.3, BS loss (BSM) |
|---|---|---|---|---|
| | | Skin Cancer, Noise=0.0 | | |
| ROC-AUC ↑ | **0.9645** | 0.9597 | 0.9532 | 0.9591 |
| ECE ↓ | 0.0418 | 0.0173 | **0.0085** | 0.0090 |
| Brier score ↓ | 0.0742 | 0.0770 | 0.0760 | **0.0667** |
| NLL ↓ | 0.3952 | 0.3734 | 0.3850 | **0.3528** |
| | | Skin Cancer, Noise=20.0 | | |
| ROC-AUC ↑ | 0.9653 | **0.9661** | 0.9576 | 0.9615 |
| ECE ↓ | 0.0392 | 0.0156 | **0.0079** | 0.0128 |
| Brier score ↓ | 0.0696 | **0.0684** | 0.0725 | 0.0690 |
| NLL ↓ | 0.3864 | **0.3442** | 0.3700 | 0.3517 |
| Breast Ultrasound, Noise=0.0 | | | | |
| ROC-AUC ↑ | 0.9924 | 0.9934 | 0.9852 | **0.9935** |
| ECE ↓ | 0.0205 | 0.0145 | 0.0078 | 0.0122 |
| Brier score ↓ | 0.0249 | **0.0223** | 0.0242 | 0.0233 |
| NLL ↓ | 0.1637 | 0.1342 | 0.1507 | **0.1331** |
| Breast Ultrasound, Noise=20.0 | | | | |
| ROC-AUC ↑ | **0.9947** | 0.9832 | 0.9922 | 0.9922 |
| ECE ↓ | **0.0222** | 0.0230 | 0.0251 | 0.0251 |
| Brier score ↓ | 0.0268 | 0.0353 | **0.0249** | **0.0249** |
| NLL ↓ | 0.1788 | 0.2103 | **0.1474** | **0.1474** |

* **No Mixup** means we only use common data augmentation during training.

### 4.3 Distance perception

In order to gain additional insights into the distance perception of various networks, as well as the influence of *Mixup* and bootstrapping strategy, we explore the correlation between uncertainty and the distance to the (small) training set. Since there is no straightforward and semantically meaningful distance between images, we use the cosine distance between the 512-dimensional representations of the features as the similarity[7, 31]. The distance of a test image $x_i$ to the training dataset is defined as $\min\{d(x_i, y_i) : y_i \in D_{train}\}$ [7]. We analyzed the domain perception ability on clinical breast ultrasound dataset. The experiment shows that the distance between validation and training set is highly correlated with the uncertainty and the BSM model shows the highest of all (Table. 4, Fig. 4. First row). Here, Spearman Correlation Coefficient[32] is used to measure the rank correlation between the two variables. Table. 4. implies that the farther away a prediction is from an observation point, the higher the uncertainty gains.

**Table 4.** Spearman Correlation Coefficient between similarity and uncertainty

| Breast | Spearman [32] | P-value |
|---|---|---|
| No Mixup, TTA:0 | -0.1590 | 0.0016 |
| No Mixup, TTA: 64 | -0.1667 | 0.0010 |
| Mixup=0.3, CE loss | -0.2631 | <<0.0001 |
| Mixup=0.3, BS loss | **-0.4428** | <<0.0001 |

In addition, we considered the influence of domain transformation on the uncertainty. We used the weights of models previously trained for breast ultrasound dataset to predict the skin cancer validation sets. As is shown in the second row of Fig. 5., we discovered that *Mixup* and BSM model generally sustained high uncertainty for domain transfer data and was able to perceive the semantic distance of data. It is highly sensitive for preventing out-of-distribution data attacking in disease diagnosis, reflecting the “cognition” to the “unknown”. Although TTA has good distance awareness for in-domain data, it is helpless for the out-of-domain.

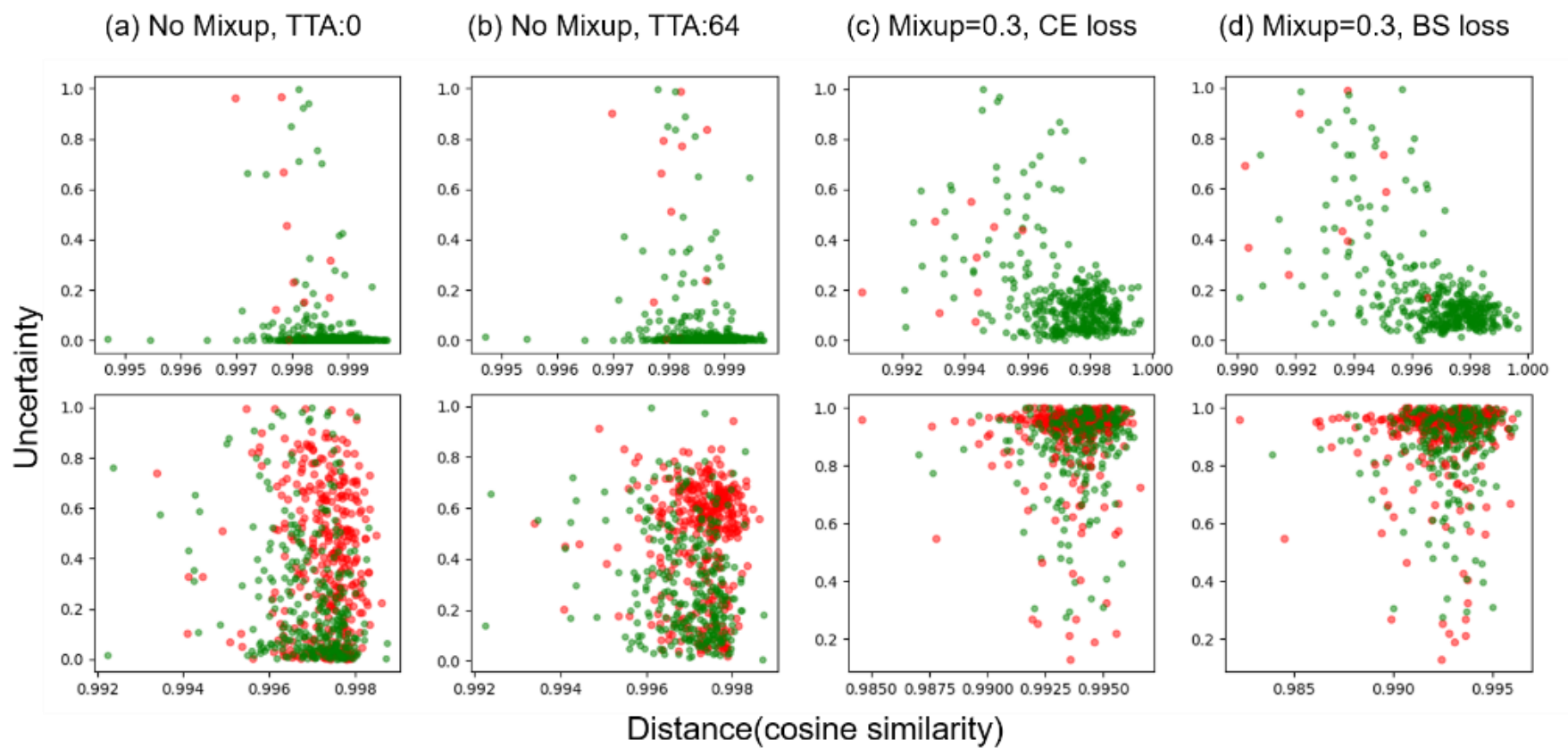

**Fig. 4. Uncertainty vs. Distance between training set (Breast ultrasound) and validation set.** The red dot indicates False prediction while the green dot indicates the True. (a) No Mixup, TTA:0. (b) No Mixup, TTA: 64. (c) Mixup=0.3, CE loss. (d) Mixup=0.3, Bootstrapping loss. The first row is all about the same datasets while the second row is about the different.

# 5 Conclusion

It is of great significance for computer-aided diagnosis to explicitly modeling uncertainty in low-data regimes. In this paper, we introduced the BS loss combined with *Mixup* that modeling data-dependent uncertainty on small clinical data sets and calibrate model to lower ECE, NLL and Brier score than conventional data augmentation ,deep ensemble and MC dropout. It provided a more reliable method to learn sensible representation and successfully exhibit much robustness to label noise. As we further illustrate, *Mixup* is more responsive to domain-variant latent distribution and the semantic distance than the TTA method. Compared with *Mixup*, the BSM model consider data noise as well as label noise, which is more profitable to noise attack and out-of-distribution detection. In summary, this paper verified the importance of data-dependent uncertainty modeling on medical images with small data sets and proposed a more reliable and low-risk BSM model for clinical diagnosis.

# ETHICS APPROVAL AND CONSENT To PARTICIPATE

The collection and analysis of subjects' breast ultrasound data in this study were approved by the Ethics Committee of Shenzhen Hospital of Peking University in China and the Shenzhen City Baoan District Women’s and Children’s Hospital in China. (Approval number: ChiCTR2100047368)

## HUMAN AND ANIMAL RIGHTS

No animals/humans were used for studies that are basis of this research.

## FUNDINGS

This research was supported by Foundation of Shenzhen Science and Technology Planning Project (No. GJHZ20200731095205015), the International Cooperation Foundation of Tsinghua Shenzhen International Graduate School (No. HW2021001), and The Natural Science Foundation of Guangdong Province (No. 2022A1515011120)

# 1 Supplementary Tables

**Table 1.** Different alpha values of *Mixup* models in skin cancer dataset

| | No *Mixup* | Alpha=0.3 | Alpha=0.5 | Alpha=0.8 | Alpha=1.0 | Alpha=32 |
|---|---|---|---|---|---|---|
| ROC-AUC | **0.9645** | 0.9532 | 0.9525 | 0.9542 | 0.9609 | 0.9530 |
| ECE | 0.0418 | **0.0085** | 0.0173 | 0.0295 | 0.0408 | 0.0276 |
| Brier score | 0.0742 | 0.0760 | 0.0768 | **0.0732** | 0.0753 | 0.0849 |
| NLL | 0.3952 | 0.3850 | 0.3959 | **0.3789** | 0.3804 | 0.4214 |

# 2 Supplementary Figures

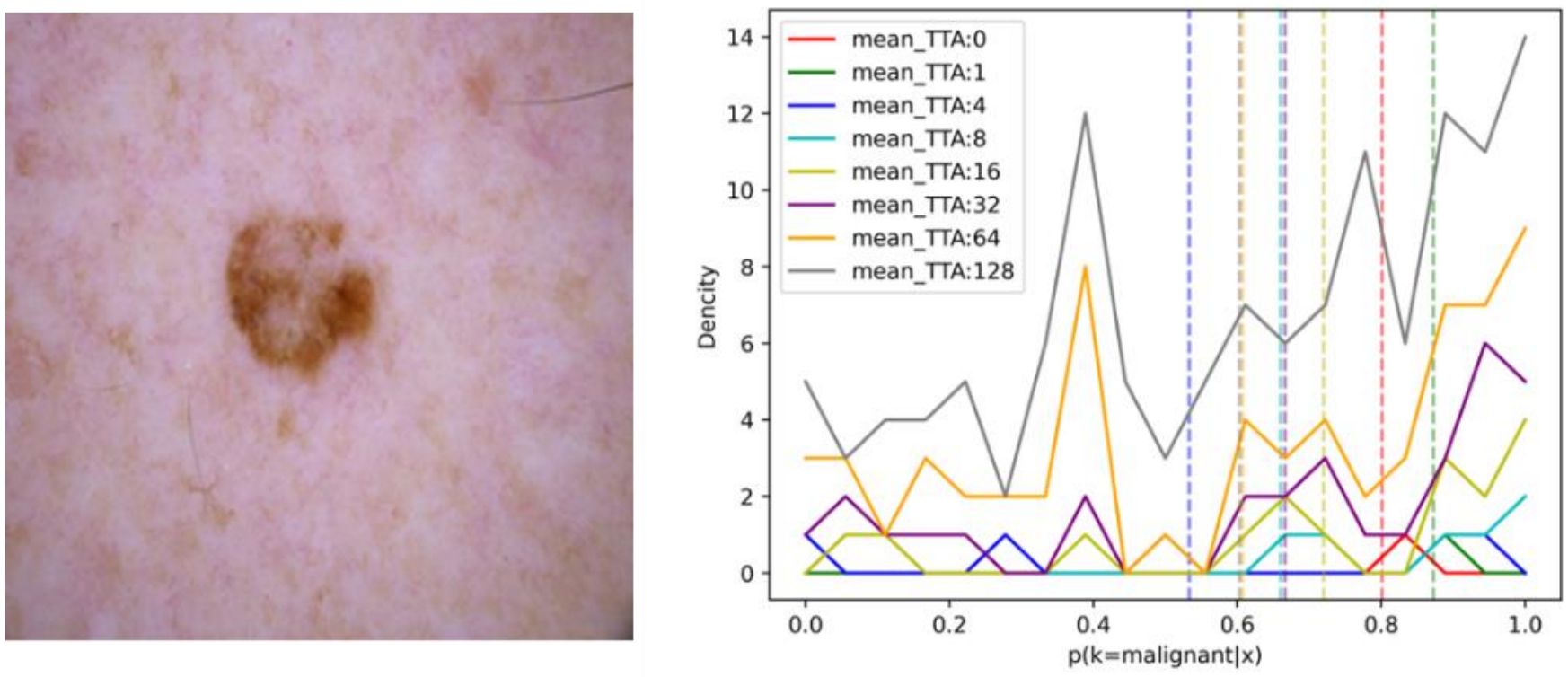

**Fig. 1.** A sample of skin cancer applied to different times of test-time augmentation. The average is indicated by the dotted line of every degree of TTA.

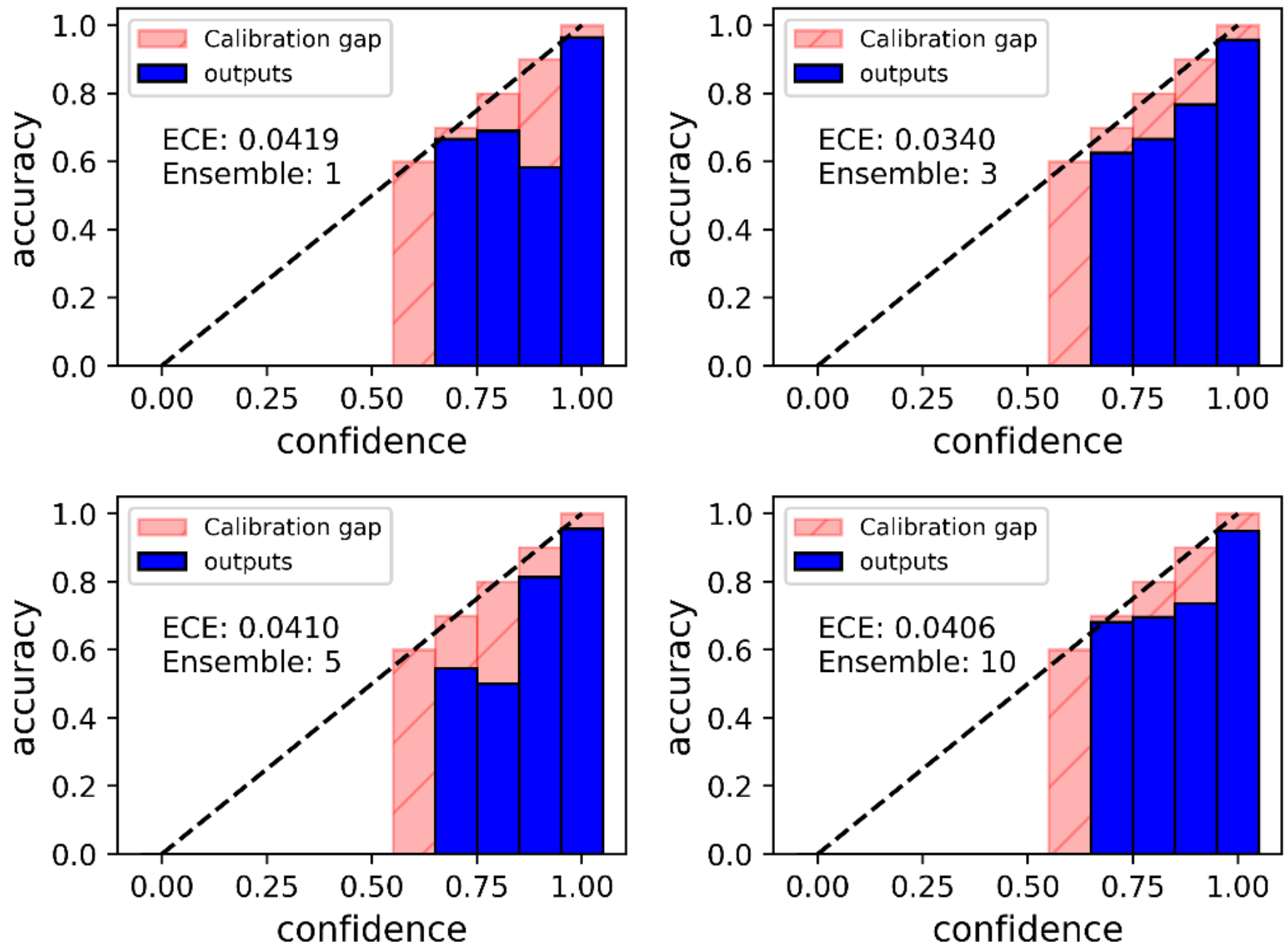

**Fig. 2.** ECE of four ensemble models(M=1, M=3, M=5, M=10). No TTA. Ensemble models are generally overconfident as $conf - acc > 0$.

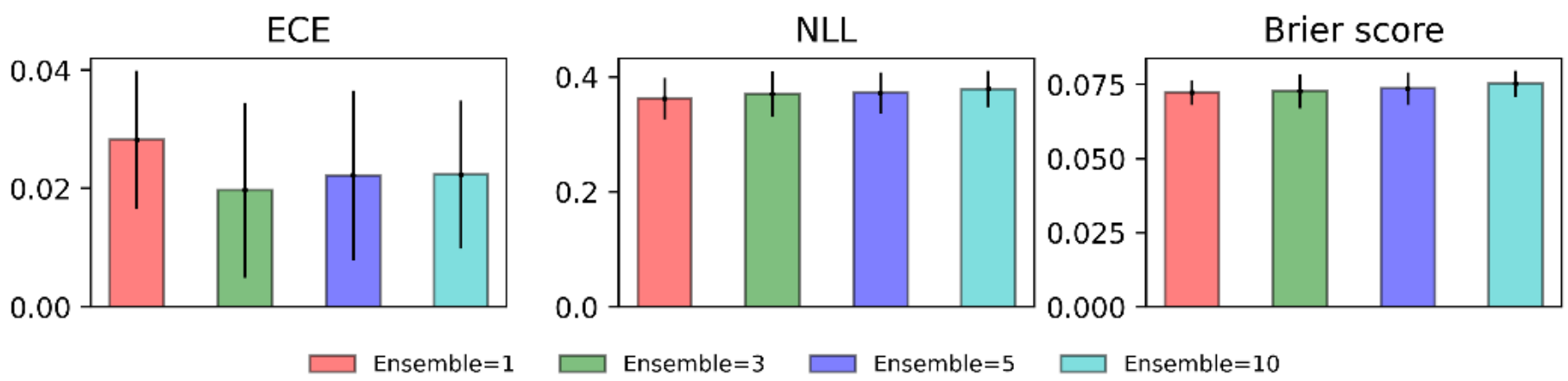

**Fig.3.** The metrics of ensemble models(M=1, M=3, M=5, M=10), where we applied different TTA degrees (TTA0, TTA1, TTA4, TTA8, TTA16, TTA32, TTA64, TTA128) and plot the error.